\documentclass{article}
\usepackage{spconf,amsmath,graphicx,hyperref}
\usepackage{booktabs}
\usepackage{tabularx}
\usepackage{array}
\usepackage{makecell}
\usepackage{amssymb} 
\usepackage[table]{xcolor}

\newcolumntype{Y}{>{\centering\arraybackslash}X}

\definecolor{bestgreen}{RGB}{218,236,218}
\definecolor{secondyellow}{RGB}{250,243,196}

\newcommand{\score}[2]{\ensuremath{#1 \pm #2}}
\newcommand{\bestscore}[2]{\cellcolor{bestgreen}\ensuremath{#1 \pm #2}}
\newcommand{\secondscore}[2]{\cellcolor{secondyellow}\ensuremath{#1 \pm #2}}
\newcommand{\bestval}[1]{\cellcolor{bestgreen}\ensuremath{#1}}
\newcommand{\secondval}[1]{\cellcolor{secondyellow}\ensuremath{#1}}

\title{HyCoSeq: Contextual Hyperbolic Representation Learning for Genomic Sequences}

\name{
Chenhao Zeng
\qquad
Zhibin Pu
\qquad
Shufei Ge\sthanks{Corresponding author, E-mail: geshf@shanghaitech.edu.cn.}}

\address{
Institute of Mathematical Sciences, ShanghaiTech University, Shanghai, China}

\begin{document}
\setlength{\skip\footins}{9pt}
\ninept

\maketitle

\begin{abstract}
Hyperbolic geometry provides a natural inductive bias for genomic representation learning, but existing hyperbolic genomic models primarily use Lorentz convolutions to learn local sequence representations, while their residual pathways do not directly aggregate full Lorentz representations. We propose HyCoSeq, a contextual hyperbolic representation learning framework for genomic sequences. HyCoSeq incorporates weighted Lorentzian residual aggregation into multi-curvature Lorentz encoding, allowing full Lorentz representations to participate directly in geometry-consistent local aggregation. It further introduces a bidirectional long short-term memory network that integrates information from both sequence directions to learn contextual relationships among local representations at different positions within a genomic sequence, thereby extending local hyperbolic convolutional encoding to sequence-level contextualized representations. Extensive experiments across diverse genomic tasks show that HyCoSeq outperforms existing hyperbolic baselines and, without large-scale genomic pretraining, achieves competitive performance against substantially larger pretrained DNA language models. Code is available at \href{https://github.com/zchhh01/Hycoseq}{Hycoseq}.
\end{abstract}


\section{INTRODUCTION}
\label{sec:introduction}

Genomic sequence classification is a key task in genomics, playing an important role in predicting protein-binding specificities, chromatin accessibility, histone marks, and the functional effects of noncoding variants~\cite{alipanahi2015predicting,zhou2015predicting,kelley2016basset}. Deep neural networks learn informative representations directly from raw DNA sequences. Diverse deep learning architectures have further advanced genomic sequence modeling across analysis tasks~\cite{quang2016danq,avsec2021effective,dalla2025nucleotide}. Despite these advances, most existing approaches still learn genomic representations in Euclidean spaces, with comparatively limited attention to representation geometry.

Recent work suggests that shared ancestry and divergence may induce latent hierarchical structure among genomic sequences~\cite{khan2025hyperbolic}. Hyperbolic spaces provide a natural inductive bias for such hierarchies: negative curvature induces exponential volume growth, enabling low-distortion embeddings of branching structures~\cite{nickel2017poincare,nickel2018learning,sala2018representation}. Hyperbolic neural networks extend core neural network operations beyond Euclidean geometry, enabling representations to be learned on hyperbolic manifolds~\cite{ganea2018hyperbolic}. Building on the fully hyperbolic convolutional neural network (CNN)~\cite{bdeir2024fully}, Khan et al. introduced Hyperbolic Genome Embeddings (HGE) for genomic sequence representation learning~\cite{khan2025hyperbolic}. Its favorable performance across diverse genomic tasks not only provides empirical support for the use of hyperbolic geometry in genomic representation learning but also highlights its potential for genomic sequence modeling.

Despite its demonstrated effectiveness, HGE still exhibits limitations in residual feature aggregation and sequence-level context modeling. The residual connection in HGE follows the design of the fully hyperbolic convolutional framework, reconstructing the timelike component after aggregating the spacelike components to satisfy the Lorentz manifold constraint~\cite{bdeir2024fully,khan2025hyperbolic}. However, the full Lorentz representations do not directly participate in residual aggregation, and this residual formulation has been noted to lack a clear geometric interpretation within hyperbolic geometry~\cite{he2025lorentzian}. At the sequence level, the Lorentz convolutional blocks primarily encode local genomic patterns within finite receptive fields. For many regulatory tasks, predictive signals depend not only on individual motifs but also on their spatial arrangement, inter-motif spacing, and flanking sequence context~\cite{avsec2021base,de2022deepstarr}. Accordingly, bidirectional sequence modeling can complement local convolutional encoding by integrating contextual information from both sequence directions~\cite{quang2016danq,luo2025hybprom}.

Motivated by these observations, we propose HyCoSeq, a contextual hyperbolic representation learning framework that jointly addresses geometry-consistent aggregation within local Lorentz encoding and bidirectional sequence-level context modeling. HyCoSeq first learns local hyperbolic representations through multi-curvature Lorentz convolution. Within each block, HyCoSeq adopts the weighted residual formulation of Lorentzian Residual Neural Networks (LResNet)~\cite{he2025lorentzian}, allowing full Lorentz representations to participate directly in residual aggregation. A bidirectional long short-term memory (BiLSTM) network~\cite{hochreiter1997long,schuster1997bidirectional} then operates on the spacelike coordinates of these local representations to model contextual dependencies across sequence positions using information from both sequence directions. The resulting contextual features are projected back onto the Lorentz manifold and processed by the Lorentz classification head. Our contributions are as follows:
\begingroup
\setlength{\parskip}{0pt}

\par
\noindent\hspace*{\parindent}(1) We propose HyCoSeq, a contextual hyperbolic representation learning framework that couples geometry-consistent local Lorentz encoding with bidirectional sequence-level context modeling for genomic sequence classification.
\par
\noindent\hspace*{\parindent}(2) We integrate a weighted Lorentzian residual connection into multi-curvature Lorentz convolution for genomic sequence modeling, enabling direct aggregation of full Lorentz representations.
\par
\noindent\hspace*{\parindent}(3) We conduct extensive experiments across diverse genomic sequence classification tasks, demonstrating the competitive performance of HyCoSeq and the effectiveness of the proposed framework.
\endgroup

\section{PRELIMINARIES}
\label{sec:preliminaries}

Our framework is formulated in the Lorentz model of hyperbolic geometry, where a $d$-dimensional hyperbolic space with constant curvature $K<0$ is represented by the upper sheet of a two-sheeted hyperboloid embedded in the $(d+1)$-dimensional Minkowski space $\mathbb{R}^{d+1}$. A point $\mathbf{x}=[x_{\mathrm{time}},\mathbf{x}_{\mathrm{space}}]\in\mathbb{R}^{d+1}$ consists of a timelike component $x_{\mathrm{time}}\in\mathbb{R}$ and a spacelike component $\mathbf{x}_{\mathrm{space}}\in\mathbb{R}^{d}$. For $\mathbf{x},\mathbf{y}\in\mathbb{R}^{d+1}$, their Lorentzian inner product is $\langle\mathbf{x},\mathbf{y}\rangle_{\mathcal L}=-x_{\mathrm{time}}y_{\mathrm{time}}+\langle\mathbf{x}_{\mathrm{space}},\mathbf{y}_{\mathrm{space}}\rangle$. The corresponding Lorentz manifold is defined as
\begin{equation}
\mathbb{L}_{K}^{d}
=
\left\{
\mathbf{x}\in\mathbb{R}^{d+1}
\,\middle|\,
\langle\mathbf{x},\mathbf{x}\rangle_{\mathcal L}
=
\frac{1}{K},
\;
x_{\mathrm{time}}>0
\right\}.
\label{eq:lorentz_manifold}
\end{equation}
Accordingly, the timelike component is determined by the spacelike component as
$x_{\mathrm{time}}=\sqrt{\|\mathbf{x}_{\mathrm{space}}\|_2^{2}-1/K}$.

\textbf{Exponential and logarithmic maps.}
For $\mathbf{x}\in\mathbb{L}_{K}^{d}$, its tangent space is
$T_{\mathbf{x}}\mathbb{L}_{K}^{d}
=
\{\mathbf{v}\in\mathbb{R}^{d+1}
\mid
\langle\mathbf{x},\mathbf{v}\rangle_{\mathcal L}=0\}$,
with
$\|\mathbf{v}\|_{\mathcal L}
=
\sqrt{\langle\mathbf{v},\mathbf{v}\rangle_{\mathcal L}}$.
For $\mathbf{v}\in T_{\mathbf{x}}\mathbb{L}_{K}^{d}$, the exponential map is defined as
\begin{equation}
\exp_{\mathbf{x}}^{K}(\mathbf{v})
=
\cosh\!\left(
\sqrt{-K}\|\mathbf{v}\|_{\mathcal L}
\right)\mathbf{x}
+
\frac{
\sinh\!\left(
\sqrt{-K}\|\mathbf{v}\|_{\mathcal L}
\right)
}{
\sqrt{-K}\|\mathbf{v}\|_{\mathcal L}
}
\mathbf{v}.
\label{eq:exp_map}
\end{equation}
Conversely, for $\mathbf{y}\in\mathbb{L}_{K}^{d}$, the logarithmic map from the manifold to the tangent space at $\mathbf{x}$ is
\begin{equation}
\log_{\mathbf{x}}^{K}(\mathbf{y})
=
\frac{
\operatorname{arcosh}\!\left(
K\langle\mathbf{x},\mathbf{y}\rangle_{\mathcal L}
\right)
}{
\sqrt{
\left(
K\langle\mathbf{x},\mathbf{y}\rangle_{\mathcal L}
\right)^2-1
}
}
\left(
\mathbf{y}
-
K\langle\mathbf{x},\mathbf{y}\rangle_{\mathcal L}\mathbf{x}
\right).
\label{eq:log_map}
\end{equation}
The origin of $\mathbb{L}_{K}^{d}$ is
$\mathbf{o}_{K}=[\sqrt{-1/K},\mathbf{0}]$.

\section{METHODOLOGY}
\label{sec:method}
\begin{figure*}[t]
    \centering
    \includegraphics[width=\textwidth]{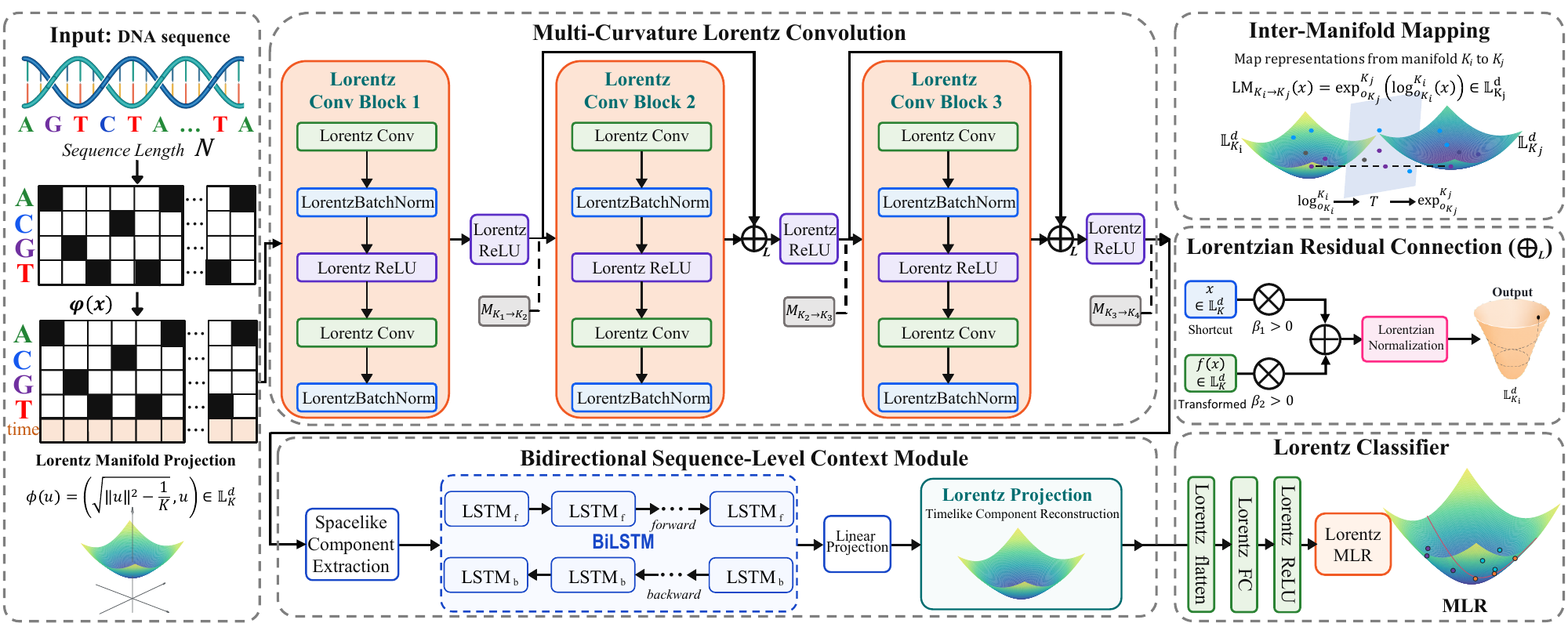}
    \caption{Overview of the HyCoSeq framework.}
    
    \label{fig:framework}
\end{figure*}

As illustrated in Fig.~\ref{fig:framework}, a nucleotide-level one-hot-encoded genomic sequence is first projected position-wise onto an initial Lorentz manifold and subsequently processed by $M$ Lorentz convolutional blocks associated with learnable curvatures $K_1,\ldots,K_M$. Within the residual blocks, weighted Lorentzian residual aggregation directly combines the full Lorentz representations of the shortcut and convolutionally transformed branches. The spacelike coordinates of the resulting local hyperbolic representations are subsequently processed by a BiLSTM. The BiLSTM outputs are first linearly transformed to the required spacelike dimensionality and then projected onto the target Lorentz manifold. Finally, the contextualized Lorentz sequence is flattened and passed through a Lorentz fully connected transformation, followed by Lorentz activation and Lorentz multinomial logistic regression (MLR) for classification.

\subsection{Multi-Curvature Lorentz Encoding}
\label{sec:lorentz_encoding}

\textbf{Multi-curvature Lorentz convolution.}
Let $\mathbf{S}=[\mathbf{s}_1,\ldots,\mathbf{s}_N]^{\top}
\in\{0,1\}^{N\times 4}$ denote the nucleotide-level one-hot encoding of a genomic sequence of length $N$, where $\mathbf{s}_{n}\in\{0,1\}^{4}$ is the one-hot vector at position $n$. For a Euclidean feature vector $\mathbf{u}\in\mathbb{R}^{q}$, its projection onto a Lorentz manifold of curvature $K<0$ is defined as
\begin{equation}
\phi_K(\mathbf{u})
=
\left[
\sqrt{\|\mathbf{u}\|_2^2-\frac{1}{K}},
\mathbf{u}
\right]
\in\mathbb{L}_K^q.
\label{eq:lorentz_projection}
\end{equation}
Applying $\phi_{K_1}$ position-wise gives the initial Lorentz sequence
$\mathbf{X}^{(0)}
=
(\phi_{K_1}(\mathbf{s}_{n}))_{n=1}^{N}
\in(\mathbb{L}_{K_1}^{4})^N$.
The projected sequence is subsequently processed by $M$ Lorentz convolutional blocks associated with learnable curvatures $K_m<0$, $m=1,\ldots,M$. The output of the $m$-th block is denoted by
$\mathbf{X}^{(m)}
=
(\mathbf{x}_{n}^{(m)})_{n=1}^{N}
\in(\mathbb{L}_{K_m}^{d})^N$.
Here, $d$ denotes the dimensionality of the spacelike coordinates, which remains fixed across the Lorentz convolutional blocks. The Lorentz convolution, normalization, and activation operators follow the fully hyperbolic CNN formulation~\cite{bdeir2024fully}, while the multi-curvature design follows HGE~\cite{khan2025hyperbolic}.

\textbf{Weighted Lorentzian residual aggregation.}
The residual operation in the fully hyperbolic CNN first aggregates the spacelike components of its input and transformed output and then reconstructs the corresponding timelike component~\cite{bdeir2024fully}. Although the resulting representation satisfies the Lorentz manifold constraint, the full Lorentz representations do not participate directly in residual aggregation. To directly aggregate the full Lorentz representations, let $\mathbf{x}\in\mathbb{L}_K^d$ be the residual-block input and let $f(\mathbf{x})\in\mathbb{L}_K^d$ be the corresponding transformed output. Following LResNet~\cite{he2025lorentzian}, the weighted Lorentzian residual connection is defined as
\begin{equation}
\mathbf{x}\oplus_{\mathcal L}f(\mathbf{x})
=
w_1\mathbf{x}+w_2f(\mathbf{x}).
\label{eq:lorentz_residual}
\end{equation}
Let $\mathbf{a}=\beta_1\mathbf{x}+\beta_2f(\mathbf{x})$ and $w_i=\beta_i/(\sqrt{-K}\sqrt{-\langle\mathbf{a},\mathbf{a}\rangle_{\mathcal L}})$ for $i\in\{1,2\}$, where $\beta_1,\beta_2>0$ are the unnormalized branch coefficients. The shared normalization factor maps the weighted combination onto $\mathbb{L}_K^d$, thereby preserving the Lorentz manifold constraint. Since $w_1$ and $w_2$ share the same normalization factor, their ratio satisfies
$w_2/w_1=\beta_2/\beta_1$.
Consequently, the residual output is invariant to any common positive rescaling of $\beta_1$ and $\beta_2$ and depends on these coefficients only through the ratio $\beta_2/\beta_1$, equivalently $w_2/w_1$. We therefore fix $\beta_1=1$ and parameterize
$\beta_2=\operatorname{softplus}(\theta)+\epsilon$,
where $\epsilon>0$ ensures strict positivity and $\theta$ is learned independently for each residual block. Under this parameterization, $\beta_2=w_2/w_1$ directly controls the relative weighting of the transformed output with respect to the block input, allowing each residual block to learn a block-specific balance between the two representations.

\textbf{Inter-manifold mapping.}
In the multi-curvature architecture, successive convolutional blocks operate on Lorentz manifolds with different learnable curvatures. A representation $\mathbf{x}\in\mathbb{L}_{K_i}^{d}$ is mapped from $\mathbb{L}_{K_i}^{d}$ to $\mathbb{L}_{K_j}^{d}$ through
\begin{equation}
\operatorname{LM}_{K_i\rightarrow K_j}(\mathbf{x})
=
\exp_{\mathbf{o}_{K_j}}^{K_j}
\left(
\log_{\mathbf{o}_{K_i}}^{K_i}(\mathbf{x})
\right)
\in\mathbb{L}_{K_j}^{d},
\label{eq:layer_mapping}
\end{equation}
where $\mathbf{o}_{K_i}$ and $\mathbf{o}_{K_j}$ denote the origins of the source and target Lorentz manifolds, respectively.

\subsection{Bidirectional Sequence-Level Context Modeling}
\label{sec:context_modeling}

Before bidirectional context modeling, the final convolutional output is mapped position-wise to $\mathbb{L}_{K_{M+1}}^{d}$ using Eq.~\eqref{eq:layer_mapping}, yielding $\widetilde{\mathbf{X}}^{(M)}=\operatorname{LM}_{K_M\rightarrow K_{M+1}}(\mathbf{X}^{(M)})\in(\mathbb{L}_{K_{M+1}}^{d})^N$. We denote its $n$-th representation by $\widetilde{\mathbf{x}}_{n}^{(M)}$. Its spacelike-coordinate matrix is $\widetilde{\mathbf{X}}_{\mathrm{space}}^{(M)}=[\widetilde{\mathbf{x}}_{1,\mathrm{space}}^{(M)},\ldots,\widetilde{\mathbf{x}}_{N,\mathrm{space}}^{(M)}]^{\top}\in\mathbb{R}^{N\times d}$.
For a given curvature $K_{M+1}$, these spacelike coordinates uniquely determine the corresponding points on the upper sheet through the Lorentz manifold constraint. We therefore perform bidirectional sequence-level context modeling on this coordinate representation to incorporate contextual information into the local hyperbolic representations.

We employ a single-layer BiLSTM with hidden size $h$ in each direction to obtain bidirectional contextual representations:
\begin{equation}
\mathbf{H}
=
\operatorname{BiLSTM}
\left(
\widetilde{\mathbf{X}}_{\mathrm{space}}^{(M)}
\right)
\in\mathbb{R}^{N\times 2h}.
\label{eq:bilstm_context}
\end{equation}
The $n$-th row of $\mathbf{H}$ is given by
$\mathbf{h}_{n}
=
[\overrightarrow{\mathbf{h}}_{n};
\overleftarrow{\mathbf{h}}_{n}]
\in\mathbb{R}^{2h}$,
where $\overrightarrow{\mathbf{h}}_{n}$ and
$\overleftarrow{\mathbf{h}}_{n}$ summarize information propagated from preceding and following sequence positions, respectively. The resulting bidirectional contextual representations are linearly transformed as
\begin{equation}
\mathbf{Z}
=
\mathbf{H}\mathbf{W}_{p}
+
\mathbf{b}_{p}
\in\mathbb{R}^{N\times d},
\label{eq:linear_projection}
\end{equation}
where $\mathbf{W}_{p}\in\mathbb{R}^{2h\times d}$ and $\mathbf{b}_{p}\in\mathbb{R}^{d}$ are learnable parameters. This transformation maps the $2h$-dimensional bidirectional contextual feature at each position to a $d$-dimensional vector, matching the dimensionality of the spacelike component in the original Lorentz representation. This alignment allows the contextual features to be projected position-wise back onto the Lorentz manifold while preserving the input dimensionality required by the original classification head.

Let $\mathbf{z}_{n}\in\mathbb{R}^{d}$ denote the $n$-th row of $\mathbf{Z}$. The contextual representations are projected position-wise back onto the final Lorentz manifold using $\phi_{K_{M+1}}$ defined in Eq.~\eqref{eq:lorentz_projection}:
\begin{equation}
\mathbf{X}_{\mathrm{ctx}}
=
\left(
\phi_{K_{M+1}}(\mathbf{z}_{n})
\right)_{n=1}^{N}
\in
\left(
\mathbb{L}_{K_{M+1}}^{d}
\right)^N.
\label{eq:context_lorentz}
\end{equation}
The contextualized Lorentz representations are subsequently processed by Lorentz flattening, a Lorentz fully connected transformation, and Lorentz activation before classification with Lorentz MLR~\cite{bdeir2024fully}. All model parameters are optimized end-to-end using the cross-entropy objective.

\section{EXPERIMENTS}

\begin{table*}[t]
\centering

\caption{
Ablation results on the seven TEB tasks, averaged over five random seeds and reported as MCC (\%, mean $\pm$ standard deviation). \textit{Lorentz Res.} denotes weighted Lorentzian residual aggregation, while \textit{Processed} and \textit{Unprocessed} denote the two pseudogene tasks. The best and second-best mean scores in each task column are highlighted in light green and light yellow, respectively.
}
\label{tab:teb_ablation}

{
\fontsize{8}{9}\selectfont

\setlength{\tabcolsep}{2.0pt}

\renewcommand{\arraystretch}{1.05}

\begin{tabular}{@{}lcc*{7}{c}@{}}

\toprule

\textbf{Model}
&
\textbf{Lorentz Res.}
&
\textbf{BiLSTM}
&
\textbf{LTR Copia}
&
\textbf{LINEs}
&
\textbf{SINEs}
&
\textbf{CMC-EnSpm}
&
\textbf{hAT-Ac}
&
\textbf{Processed}
&
\textbf{Unprocessed}
\\

\midrule

Euclidean CNN
&
\multicolumn{2}{c}{\textit{n/a}}
&
\score{54.73}{1.45}
&
\score{70.63}{1.24}
&
\score{85.15}{1.64}
&
\score{72.18}{0.32}
&
\score{87.45}{0.90}
&
\score{60.66}{0.82}
&
\score{51.94}{2.69}
\\

HGE (HCNN-M)
&
--
&
--
&
\score{68.05}{2.80}
&
\score{77.10}{2.92}
&
\score{81.85}{2.95}
&
\score{80.65}{1.30}
&
\score{91.04}{1.58}
&
\score{65.41}{5.54}
&
\score{58.36}{1.80}
\\

\addlinespace[1.5pt]

Residual variant
&
$\checkmark$
&
--
&
\secondscore{71.92}{1.23}
&
\score{81.04}{2.27}
&
\secondscore{91.84}{1.12}
&
\score{83.61}{1.16}
&
\score{91.08}{0.53}
&
\secondscore{70.27}{2.84}
&
\secondscore{61.14}{2.15}
\\

Context variant
&
--
&
$\checkmark$
&
\score{70.36}{1.79}
&
\secondscore{81.06}{1.90}
&
\score{91.81}{0.93}
&
\secondscore{83.98}{0.51}
&
\secondscore{92.48}{0.77}
&
\score{69.18}{0.85}
&
\score{58.64}{0.49}
\\

\addlinespace[1.5pt]

\textbf{HyCoSeq}
&
$\checkmark$
&
$\checkmark$
&
\bestscore{72.88}{0.92}
&
\bestscore{83.54}{0.52}
&
\bestscore{93.33}{1.12}
&
\bestscore{84.45}{0.92}
&
\bestscore{93.73}{0.36}
&
\bestscore{70.65}{0.79}
&
\bestscore{62.62}{0.97}
\\

\bottomrule

\end{tabular}
}

\end{table*}

\begin{table*}[t]
\centering

\caption{
Comparison of HyCoSeq with pretrained DNA language models and HGE baselines on 17 representative GUE datasets. Columns 2--18 report MCC (\%) on the 17 datasets. The last three columns report trainable parameter counts in millions (M), average rank across the 17 datasets (lower is better), and the number of dataset-level best results, respectively. The best and second-best dataset scores are highlighted in light green and light yellow, respectively.
}

\label{tab:gue_17datasets}

\fontsize{8}{9}\selectfont
\setlength{\tabcolsep}{0.3pt}
\renewcommand{\arraystretch}{1.04}

\begin{tabularx}{\textwidth}{
@{}
>{\raggedright\arraybackslash}p{24.8mm}
*{17}{Y}
>{\centering\arraybackslash}p{9mm}
>{\centering\arraybackslash}p{7mm}
>{\centering\arraybackslash}p{6.1mm}
@{}
}

\toprule

\textbf{Model}
&
\makecell{\textbf{H3K4}\\\textbf{me1}}
&
\makecell{\textbf{H3K4}\\\textbf{me2}}
&
\makecell{\textbf{H3K4}\\\textbf{me3}}
&
\makecell{\textbf{H3K7}\\\textbf{9me3}}
&
\makecell{\textbf{H3K9}\\\textbf{ac}}
&
\makecell{\textbf{H4}\\\textbf{ac}}
&
\makecell{\textbf{P-}\\\textbf{TATA}}
&
\textbf{hTF1}
&
\textbf{hTF2}
&
\textbf{hTF3}
&
\textbf{hTF4}
&
\makecell{\textbf{CP-}\\\textbf{TATA}}
&
\textbf{mTF1}
&
\textbf{mTF2}
&
\textbf{mTF3}
&
\textbf{mTF4}
&
\textbf{Splice}
&
\makecell{\textbf{Param}\\\textbf{(M)}}
&
\makecell{\textbf{Avg.}\\\textbf{Rank}}
&
\textbf{Wins}
\\

\midrule

Caduceus-Ph~\cite{schiff2024caduceus}
&
37.76 & 28.16 & 24.40 & 60.31 & 52.70 & 40.90
& 66.07
& 72.10 & 58.92 & 54.85 & 69.45
& 72.94
& 80.31 & 75.89 & 73.47 & 47.98
& 81.59
& 7.7
& 7.00
& 0
\\

HyenaDNA~\cite{nguyen2023hyenadna}
&
35.83 & 25.81 & 23.15 & 54.09 & 50.84 & 38.44
& 5.34
& 67.86 & 46.85 & 41.78 & 61.23
& 72.87
& 80.50 & 65.34 & 54.20 & 19.17
& 72.67
& 28.2
& 10.65
& 0
\\

DNABERT-6mer~\cite{ji2021dnabert}
&
41.44 & 32.27 & 27.81 & 61.17 & 51.22 & 37.43
& 61.56
& 70.14 & 61.03 & 51.89 & 70.97
& 76.06
& 78.94 & 71.44 & 44.89 & 42.48
& 84.07
& 89
& 7.24
& 0
\\

NT-500M-human~\cite{dalla2025nucleotide}
&
37.15 & 30.87 & 24.06 & 58.35 & 45.81 & 33.74
& 78.07
& 66.75 & 53.58 & 42.95 & 60.81
& 71.34
& 75.04 & 61.67 & 29.17 & 29.27
& 79.71
& 500
& 10.94
& 0
\\

NT-500M-1000g~\cite{dalla2025nucleotide}
&
40.45 & 31.05 & 26.16 & 59.33 & 49.29 & 36.79
& 78.23
& 70.17 & 52.73 & 45.24 & 62.82
& 73.52
& 75.49 & 64.70 & 33.07 & 34.01
& 80.97
& 500
& 9.24
& 0
\\

NT-2.5B-1000g~\cite{dalla2025nucleotide}
&
43.10 & 30.28 & 30.87 & 61.20 & 52.36 & 41.46
& 75.80
& 68.30 & 58.70 & 49.08 & 67.59
& 69.66
& 80.02 & 70.14 & 42.25 & 43.40
& 85.78
& 2500
& 7.65
& 0
\\

NT-2.5B-multi~\cite{dalla2025nucleotide}
&
\bestval{55.30}
& 36.49
& 40.34
& 64.70
& 56.01
& 49.13
& 79.43
& 70.28
& 58.72
& 51.65
& 69.34
& 72.97
& 83.76
& 71.52
& 69.44
& 47.07
& \bestval{89.35}
& 2500
& 4.71
& 2
\\

DNABERT-2~\cite{zhou2024dnabert}
&
50.52
& 31.13
& 36.27
& \bestval{67.39}
& 55.63
& 50.43
& 71.59
& \bestval{76.06}
& 66.52
& \bestval{58.54}
& \bestval{77.43}
& 74.17
& \secondval{84.77}
& \secondval{79.32}
& 66.47
& \bestval{52.66}
& 84.99
& 117
& 3.59
& 5
\\

DNABERT-2-PT~\cite{zhou2024dnabert}
&
\secondval{53.00}
& 39.89
& 41.20
& \secondval{65.46}
& \bestval{57.07}
& 50.35
& 68.79
& 71.87
& 62.96
& \secondval{55.35}
& 74.94
& 76.18
& \bestval{86.28}
& \bestval{81.28}
& 73.49
& \secondval{50.80}
& \secondval{85.93}
& 117
& 3.06
& 3
\\

HCNN-S~\cite{khan2025hyperbolic}
&
41.86
& \secondval{43.88}
& \secondval{50.58}
& 64.62
& 54.09
& \bestval{52.94}
& \secondval{82.70}
& 69.39
& \secondval{73.80}
& 44.08
& 68.43
& \secondval{81.34}
& 79.26
& 77.86
& \secondval{73.51}
& 41.27
& 81.96
& 4.6
& 4.82
& 1
\\

HCNN-M~\cite{khan2025hyperbolic}
&
39.78
& 31.27
& 33.59
& 63.35
& 52.25
& \secondval{51.86}
& 79.80
& 68.48
& 71.40
& 43.66
& 70.01
& \bestval{82.07}
& 77.41
& 77.51
& 69.73
& 43.62
& 82.23
& 4.6
& 5.94
& 1
\\

\textbf{HyCoSeq}
&
42.98
& \bestval{44.32}
& \bestval{50.87}
& 65.11
& \secondval{56.85}
& 51.78
& \bestval{82.89}
& \secondval{75.72}
& \bestval{75.07}
& 47.69
& \secondval{76.20}
& 78.67
& 77.87
& 79.29
& \bestval{77.02}
& 45.46
& 84.47
& 4.6
& 3.18
& 5
\\

\bottomrule
\end{tabularx}
\end{table*}

\begin{table}[t]
\centering
\caption{
Mean model performance (MCC) aggregated by genomic task category on GUE and GB (\%, mean $\pm$ standard error). TF and OCR denote transcription factor and open chromatin regions, respectively. The best and second-best results for each task row are highlighted in light green and light yellow, respectively.
}

\label{tab:gue_gb}

{
\fontsize{8}{9}\selectfont
\setlength{\tabcolsep}{0.8pt}
\renewcommand{\arraystretch}{1.05}

\newcommand{\aggscore}[2]{%
  \mbox{\ensuremath{#1\!\pm\!#2}}%
}

\newcommand{\aggbest}[2]{%
  \cellcolor{bestgreen}%
  \mbox{\ensuremath{#1\!\pm\!#2}}%
}

\newcommand{\aggsecond}[2]{%
  \cellcolor{secondyellow}%
  \mbox{\ensuremath{#1\!\pm\!#2}}%
}

\begin{tabular}{@{}
>{\raggedright\arraybackslash}p{20mm}
>{\centering\arraybackslash}p{15.5mm}
>{\centering\arraybackslash}p{15.5mm}
>{\centering\arraybackslash}p{15.5mm}
>{\centering\arraybackslash}p{16.5mm}
@{}}

\toprule

\textbf{Task}
&
Eucl. CNN
&
HCNN-S
&
HCNN-M
&
\textbf{HyCoSeq}
\\

\midrule

\multicolumn{5}{@{}l}{\textbf{GUE}}
\\[-1pt]

Epigenetic Marks
&
\aggscore{40.76}{4.07}
&
\aggsecond{55.31}{2.64}
&
\aggscore{48.18}{2.81}
&
\aggbest{55.82}{3.62}
\\

Human TF
&
\aggscore{52.52}{3.63}
&
\aggscore{61.12}{3.12}
&
\aggsecond{61.25}{2.86}
&
\aggbest{64.72}{4.62}
\\

Splice Site
&
\aggscore{78.64}{0.19}
&
\aggscore{80.32}{0.55}
&
\aggsecond{80.76}{0.47}
&
\aggbest{84.03}{0.24}
\\

Mouse TF
&
\aggscore{45.79}{4.72}
&
\aggsecond{61.93}{5.52}
&
\aggscore{61.52}{5.08}
&
\aggbest{64.92}{6.08}
\\

Core Promoter
&
\aggscore{70.13}{2.06}
&
\aggscore{70.12}{3.48}
&
\aggsecond{70.99}{2.39}
&
\aggbest{71.26}{3.87}
\\

Promoter
&
\aggsecond{85.80}{1.75}
&
\aggscore{85.73}{1.37}
&
\aggscore{85.66}{1.82}
&
\aggbest{85.93}{2.46}
\\

\midrule

\multicolumn{5}{@{}l}{\textbf{GB}}
\\[-1pt]

Demo
&
\aggscore{82.52}{2.79}
&
\aggscore{86.34}{2.83}
&
\aggsecond{86.48}{2.83}
&
\aggbest{87.57}{3.45}
\\

Enhancers
&
\aggsecond{39.41}{9.00}
&
\aggscore{34.18}{7.46}
&
\aggscore{28.77}{2.83}
&
\aggbest{48.47}{11.00}
\\

Regulatory
&
\aggscore{77.36}{4.73}
&
\aggsecond{86.74}{1.35}
&
\aggscore{85.05}{2.34}
&
\aggbest{89.26}{1.34}
\\

OCR
&
\aggscore{39.92}{0.38}
&
\aggsecond{56.22}{0.13}
&
\aggscore{55.36}{1.23}
&
\aggbest{56.68}{0.25}
\\

\bottomrule

\end{tabular}
}

\end{table}

\subsection{Experimental Setup}

\textbf{Datasets.}
We evaluate HyCoSeq on three genomic benchmarks: the Transposable Elements Benchmark (TEB)~\cite{khan2025hyperbolic}, the Genome Understanding Evaluation (GUE)~\cite{zhou2024dnabert}, and Genomic Benchmarks (GB)~\cite{grevsova2023genomic}. TEB covers multi-species classification of retrotransposons, DNA transposons, and pseudogenes. GUE and GB cover diverse regulatory and sequence classification tasks, including transcription-factor binding, epigenetic marks, promoters, splice sites, enhancers, and open chromatin regions. We use Matthews correlation coefficient (MCC) for all datasets.

\textbf{Implementation details.}
For a fair comparison, we follow the same data preprocessing and splits as HGE~\cite{khan2025hyperbolic}, and use the same three-block Lorentz convolutional backbone to encode local genomic patterns. We use RiemannianAdam~\cite{becigneul2019riemannian,kochurov2020geoopt} and train our models for 100 epochs with a batch size of 100 and a backbone weight decay of 0.1. The learning rates are set to $10^{-4}$, $10^{-4}$, and $10^{-5}$ for TEB, GUE, and GB, respectively, with a manifold learning rate of $2\times10^{-2}$. All learnable Lorentz curvatures are initialized to $K=-1$. For weighted Lorentzian residual aggregation, $\beta_1$ is fixed to 1, while $\beta_2$ is learnable and initialized to 1. The BiLSTM consists of a single layer with a hidden size of 16 per direction. Experiments were conducted on NVIDIA A100 GPUs with 40 GB of memory.

\subsection{Results and Discussion}
Table~\ref{tab:teb_ablation} presents the ablation results across the seven TEB tasks. Applying either weighted Lorentzian residual aggregation or bidirectional sequence-level context modeling individually improves upon HGE (HCNN-M) across all tasks, indicating that each design contributes to the performance gains. HyCoSeq combines both designs, achieves the highest mean MCC on every task, and improves the overall average MCC over HGE by 5.53 percentage points. These results support the complementary roles of the two designs: the former allows full Lorentz representations to participate directly in residual aggregation within the convolutional blocks, whereas the latter supplements local encoding by integrating contextual information from both sequence directions.

We further compare HyCoSeq and the HGE baselines with a range of pretrained DNA language models on 17 representative GUE datasets, including DNABERT~\cite{ji2021dnabert}, HyenaDNA~\cite{nguyen2023hyenadna}, Caduceus-Ph~\cite{schiff2024caduceus}, Nucleotide Transformer~\cite{dalla2025nucleotide}, and DNABERT-2~\cite{zhou2024dnabert}, to assess whether HyCoSeq remains competitive against substantially larger pretrained models. Table~\ref{tab:gue_17datasets} reports the classification performance and number of trainable parameters for each model configuration. Following the reporting protocol of HGE~\cite{khan2025hyperbolic}, the best classification performance is reported for HyCoSeq and the HGE baselines. For HyenaDNA and Caduceus-Ph, we report benchmark evaluations of these two models, whereas the results for the remaining pretrained models follow those reported by Zhou et al.~\cite{zhou2024dnabert}. Unlike these pretrained models, HyCoSeq does not rely on large-scale genomic pretraining and is trained end-to-end on each downstream dataset. HyCoSeq has approximately 4.6M trainable parameters, only 0.16\% more than HCNN-M. In contrast, the pretrained models have parameters ranging from 7.7M to 2.5B. Despite these substantial differences in model scale and training paradigm, HyCoSeq achieves the best performance on multiple GUE datasets and an average rank of 3.18, ranking second overall behind only DNABERT-2-PT, which has an average rank of 3.06 and uses about 25.4 times as many parameters as HyCoSeq. It also outperforms the hyperbolic HGE baselines on the vast majority of the reported datasets, indicating that the improvements are not confined to a particular biological task type. These results indicate that the performance gains of HyCoSeq are not primarily attributable to increased model capacity. By integrating geometry-consistent residual aggregation with sequence-level contextualization, HyCoSeq remains competitive with substantially larger pretrained genomic models while retaining a compact hyperbolic architecture.

Table~\ref{tab:gue_gb} further evaluates HyCoSeq at the task-category level. HyCoSeq achieves the highest mean MCC across all reported task groups on both GUE and GB, spanning epigenetic-mark prediction, transcription-factor binding prediction, splice-site prediction, promoter detection, and the classification of enhancers, regulatory elements, and open chromatin regions. Although the magnitude of the gains varies across task groups, HyCoSeq maintains an overall performance advantage. Together with the preceding comparisons, these findings indicate that the benefits of the proposed representation-learning framework are not confined to a particular benchmark or task category, supporting its consistent effectiveness across diverse genomic classification settings.

\section{CONCLUSION}
\label{sec:conclusion}
We presented HyCoSeq, a contextual hyperbolic representation learning framework for genomic sequence classification. By combining multi-curvature Lorentz encoding, geometry-consistent residual aggregation, and bidirectional sequence-level contextualization, HyCoSeq enhances local hyperbolic representation learning and sequence-level contextual modeling. Experiments across diverse genomic classification settings show that HyCoSeq outperforms existing hyperbolic baselines and remains competitive with substantially larger pretrained DNA language models using fewer parameters. These results demonstrate the effectiveness of HyCoSeq and highlight the potential of hyperbolic representation learning for genomic sequence modeling.
\vfill\pagebreak
\section{Acknowledgments}
This work was supported by the National Natural Science Foundation of China (12401383) and the startup fund of ShanghaiTech University. The authors thank ShanghaiTech University for providing access to its HPC platform.
\bibliographystyle{IEEEbib}
\bibliography{strings,refs}

\end{document}